\documentclass{article}

\PassOptionsToPackage{round}{natbib}

\usepackage[dblblindworkshop, final]{neurips_2025}

\usepackage[utf8]{inputenc}       
\usepackage[T1]{fontenc}          

\usepackage{booktabs,multirow,tabularx} 
\usepackage{ragged2e}             
\usepackage{enumitem}             
\usepackage{amsfonts,amsmath,amssymb,mathtools,physics} 
\usepackage{dsfont}               
\usepackage{nicefrac}             
\usepackage{microtype}            
\usepackage[dvipsnames,table]{xcolor} 

\usepackage[hidelinks]{hyperref}  
\hypersetup{
    colorlinks=true,
    urlcolor=Blue,
    linkcolor=Blue,
    citecolor=Blue,
}
\usepackage[nameinlink]{cleveref} 
\crefname{figure}{figure}{figures}
\crefname{Figure}{Figure}{Figures}
\usepackage{url}                  

\usepackage{algorithm}            
\usepackage[endLComment=,italicComments=false]{algpseudocodex}
\algrenewcommand\algorithmicindent{3.5mm}

\usepackage{subcaption}           
\usepackage{subfiles}             

\usepackage{adjustbox}

\usepackage{tcolorbox}
\tcbuselibrary{listings,breakable}

\newtcolorbox[
    auto counter,
    list inside=prompt,
    crefname={prompt}{prompts},
]{prompt}[2][]{
    title={Prompt \thetcbcounter: #2},
    colbacktitle=black!60,
    coltitle=white,
    colframe=black!60,
    colback=lightgray!30,
    boxsep=5pt,
    left=3pt,
    right=3pt,
    top=3pt,
    bottom=3pt,
    boxrule=0pt,
    breakable,
    #1,
}

\DeclareMathOperator*{\argmax}{\mathrm{argmax}}

\DeclareMathOperator*{\minimize}{\mathrm{minimize}}

\newcommand{\E}{\ensuremath{\mathbb{E}}}

\newcommand{\TERM}{\ensuremath{\mathtt{TERM}}}
\newcommand{\STOP}{\ensuremath{\mathtt{STOP}}}
\newcommand{\CONT}{\ensuremath{\mathtt{CONT}}}
\newcommand{\llb}{\ensuremath{[\![}}
\newcommand{\rrb}{\ensuremath{]\!]}}

\title{Stop-RAG: Value-Based Retrieval Control \\ for Iterative RAG}

\author{%
    Jaewan Park\thanks{Equal contribution} \\
    Seoul National University \\
    \href{mailto:jaejae1112@snu.ac.kr}{\color{black}\texttt{jaejae1112@snu.ac.kr}}
    \And
    Solbee Cho\footnotemark[1] \\
    Seoul National University \\
    \href{mailto:sbcho0325@snu.ac.kr}{\color{black}\texttt{sbcho0325@snu.ac.kr}}
    \And
    Jay-Yoon Lee\thanks{Corresponding author} \\
    Seoul National University \\
    \href{mailto:lee.jayyoon@snu.ac.kr}{\color{black}\texttt{lee.jayyoon@snu.ac.kr}}
}

\begin{document}

\maketitle

\begin{abstract}
    Iterative retrieval-augmented generation (RAG) enables large language models to answer complex multi-hop questions, but each additional loop increases latency, costs, and the risk of introducing distracting evidence, motivating the need for an efficient stopping strategy. Existing methods either use a predetermined number of iterations or rely on confidence proxies that poorly reflect whether more retrieval will actually help. We cast iterative RAG as a finite-horizon Markov decision process and introduce \textbf{Stop-RAG}, a value-based controller that adaptively decides when to stop retrieving. Trained with full-width forward-view Q($\lambda$) targets from complete trajectories, Stop-RAG learns effective stopping policies while remaining compatible with black-box APIs and existing pipelines. On multi-hop question-answering benchmarks, Stop-RAG consistently outperforms both fixed-iteration baselines and prompting-based stopping with LLMs. These results highlight adaptive stopping as a key missing component in current agentic systems, and demonstrate that value-based control can improve the accuracy of RAG systems. Our code is available at \url{https://github.com/chosolbee/Stop-RAG}.
\end{abstract}

\section{Introduction}
\label{sec:intro}
The rise of agentic artificial intelligence (AI) \citep{aiagents} is rapidly reshaping how large language models (LLMs) are conceived and deployed. Recent research increasingly views LLMs not as standalone responders in single-turn interactions, but as components within broader agentic systems. A central goal in this direction has been to equip LLMs with the ability for \emph{multi-turn tool use}, enabling them to iteratively query external resources, call specialized models, and refine their outputs \citep{kimik2,glm4.5}. However, comparatively little attention has been given to the skill of adaptive decision-making, specifically, the ability to judge when further tool calls or further generation are unnecessary and the task can be completed with the information already gathered. This gap highlights a critical yet underexplored dimension in building effective AI systems.

Existing approaches often sidestep this challenge by treating finishing as just another tool call, expecting large-scale training to implicitly teach models when to stop \citep{react,reflexion,artist}. However, making this decision fundamentally requires the model to self-reflect and evaluate its own past reasoning trace. As \citet{rlvmr} observe, most current agents struggle here, leading to long, inefficient sequences of actions.

The challenge of deciding when to stop becomes even more critical in the context of retrieval-augmented generation (RAG) \citep{retro,atlas,replug}. Recent RAG systems have also shifted towards employing iterative retrieve-generate loops to solve complex questions \citep{selfask,ircot,iterretgen}. In deployed systems, however, each additional iteration carries costs---latency, tokens, and the risk of distractors---while benefits vary across queries, making optimal stopping crucial \citep{gsm-ic,retrobust}. Hence, deciding when to stop these iterations fundamentally shapes system performance in iterative RAG.

However, existing stopping mechanisms in iterative RAG yet face critical limitations. Some approaches simply rely on LLM's self-assessment \citep{ircot,searcho1,iterdrag} or internal telemetry signals \citep{flare,dragin}, but both have been shown to be unreliable \citep{knowledgeboundary,overconfidence}. More sophisticated methods train specialized modules \citep{selfrag,probingrag}, yet their supervision targets are shaped only to reflect the expected answer quality if stopped at the current step. This present-focused training can be misleading: for instance, even if all necessary evidence is already retrieved, the presence of many irrelevant documents may yield a low score and trigger further retrieval, which only adds more noise and harms the final answer. Conversely, an apparently confident answer at an early step might tempt the system to stop, even when continuing would clearly improve completeness.

To address these problems, we reframe iterative RAG as a finite-horizon Markov decision process (MDP) and train a Q-network using Q($\lambda$), a Q-learning variant with $\lambda$-returns. This approach yields forward-looking estimates of whether continuing retrieval will improve answer quality. By estimating and comparing both immediate and future gains, our method enables more reliable stopping decisions. Importantly, this does not require internal telemetry and integrates with existing pipelines, making it compatible with black-box LLMs and widely deployable as a modular component.

We summarize our contributions as follows:
\begin{itemize}[leftmargin=0.6cm]
    \item We identify the underexplored problem of adaptive stopping in agentic AI, especially within iterative RAG.
    \item We propose a novel formulation of iterative RAG as a finite-horizon MDP and train a Q-network using Q($\lambda$) to provide forward-looking estimates of stopping quality.
    \item We demonstrate that our approach improves performance on multi-hop question-answering (QA) benchmarks, while remaining modular and compatible with black-box LLMs and existing iterative RAG pipelines.
\end{itemize}

\section{Related work}
\label{sec:related_work}
\paragraph{Iterative RAG}
Iterative RAG systems tackle complex queries requiring multi-step reasoning by running repeated retrieve-generate loops that incrementally accumulate evidence. Compared to single-shot retrieval, this paradigm (1) decomposes complex problems into tractable sub-questions, (2) continuously refines retrieval queries using intermediate hypotheses and findings, and (3) enables self-correction when early retrievals are insufficient. CoRAG \citep{corag} explicitly learns step-wise retrieval via automatically generated intermediate chains, while Iter-RetGen \citep{iterretgen} uses prior model outputs to guide subsequent retrieval steps. However, these methods rely on a fixed number of iterations, which leads to fundamental inefficiencies regarding query-specific complexity: simple questions may waste computation through unnecessary loops, whereas difficult ones may halt prematurely before enough evidence is gathered. These limitations motivate adaptive approaches that dynamically decide when to stop.

\paragraph{Adaptive RAG}
Adaptive RAG methods address the limitations of fixed iteration budgets by dynamically determining when to stop the retrieval process based on query-specific characteristics and intermediate results. Existing approaches in adaptive RAG can be categorized into three main paradigms based on how they make stopping decisions.
\textbf{Prompting-based} methods allow models to textually decide whether to continue or stop retrieval. IRCoT \citep{ircot} and IterDRAG \citep{iterdrag} interleave reasoning and retrieval, stopping when a final answer is produced. Search-o1 \citep{searcho1} uses special search tokens during reasoning and terminates upon a final answer. 
\textbf{Confidence-based} methods leverage uncertainty signals to trigger stopping decisions. FLARE \citep{flare} uses token-level probability thresholds, while DRAGIN \citep{dragin} combines multiple uncertainty measures including entropy and attention weights to determine when retrieval is necessary.
\textbf{Training-involved} methods learn explicit stopping mechanisms. Self-RAG \citep{selfrag} trains a model to generate reflection tokens at each step to guide stopping decisions, using data distilled from a larger LLM. Probing-RAG \citep{probingrag} trains a prober to score if future retrieval is necessary, with supervision from binary labels indicating whether the generated answer at the current step is correct. Adaptive-RAG \citep{adaptiverag} instead trains a complexity classifier, but operates only pre-execution and cannot be applied within the iterative loop.
Unlike these methods, our approach does not rely on model-internal or present-focused signals, but instead provides forward-looking value estimates by framing stopping as a decision in a finite-horizon MDP.

\section{Preliminaries}
\label{sec:preliminaries}
\paragraph{Q-learning}
In our work, we formulate the iterative RAG setup as a finite-horizon MDP \citep{suttonbartoed1} and apply reinforcement learning (RL) to optimize the adaptive retrieval strategy. An MDP $\mathcal{M}$ is defined as a tuple $\qty(\mathcal{S}, \mathcal{A}, r, p)$ where $\mathcal{S}$ is the state space, $\mathcal{A}$ is the action space, $r: \mathcal{S} \times \mathcal{A} \rightarrow \mathbb{R}$ is the reward function, and $p: \mathcal{S} \times \mathcal{A} \rightarrow \Delta\qty(\mathcal{S})$ is the function that maps a state-action pair to its corresponding transition dynamics distribution. We use the notation $p\qty(s' \mid s, a)$ to denote the probability of transitioning to state $s'$ when action $a$ is taken in state $s$. A policy $\pi: \mathcal{S} \rightarrow \Delta\qty(\mathcal{A})$ is a function that maps a state to a distribution over actions. In value-based RL, specifically Q-learning \citep{watkins1989qlearning,mnih2013dqn}, we take a two-step approach where we first iteratively approximate the optimal action-value function
\begin{equation}
    Q^*\qty(s_t, a_t) = \max_\pi\E_{p, \pi}\qty[\sum_{k=0}^{T-t-1}\gamma^{k}r\qty(s_{t+k}, a_{t+k}) \Biggm\vert s_t, a_t],
\end{equation}
\footnote{The subscript $p, \pi$ is an abbreviation for $s_{t+1} \!\sim\! p\qty(\cdot \mid s_t, a_t), \, a_{t+1} \!\sim\! \pi\qty(\cdot \mid s_{t+1}), \, \ldots, \, a_{T-1} \!\sim\! \pi\qty(\cdot \mid s_{T-1})$.}
then derive a greedy policy $\pi^*\qty(s) = \argmax_{a}Q^*\qty(s, a)$. $\pi^*$ provably converges to the true optimal policy under mild assumptions \citep{watkins1992qlearning,jaakola1994qlearning,tsitsiklis1994qlearning}.

\paragraph{Full-width Q($\boldsymbol{\lambda}$) for finite-horizon MDPs}
Function approximation-based Q-learning in its most basic form iteratively performs the following optimization process on an off-policy dataset $\mathcal{D}$:
\begin{equation} \label{eq:qlearning}
    \minimize_{\theta}\E_{\qty(s, a) \sim \mathcal{D}}\Biggl[\biggl(\underbrace{r\qty(s, a) + \gamma\E_{s' \sim p\qty(\cdot \mid s, a)}\qty[\max_{a' \in \mathcal{A}}\llb Q_\theta\qty(s', a')\rrb \biggm\vert s, a]}_{\eqcolon \hat{Q}\qty(s, a)} - Q_\theta\qty(s, a)\biggr)^2\Biggr].
\end{equation}
\footnote{$\llb \cdot \rrb$ denotes the stop-gradient operator.}
As this one-step bootstrap target $\hat{Q}$ is biased, we can unroll the backup over $n$ steps, reducing bias at the expense of higher variance. This yields the following recursive definition of the $n$-step Bellman optimality backup:
\begin{align}
    \hat{Q}^{(0)}\qty(s, a) &= \llb Q_\theta\qty(s, a)\rrb, \label{eq:q_0} \\
    \hat{Q}^{(n)}\qty(s, a) &= r\qty(s, a) + \gamma\E_{s' \sim p\qty(\cdot \mid s, a)}\qty[\max_{a' \in \mathcal{A}}\hat{Q}^{(n-1)}\qty(s', a') \biggm\vert s, a]. \qquad \qty(n > 0) \label{eq:q_n_full}
\end{align}
Normally, the expectation over $s'$ and maximization over $a'$ up to the $\qty(n-1)$-th transition are replaced by sampling and integrated into $\mathcal{D}$. However, if the action space is sufficiently small, at each data collection step we can instead build a \emph{full-width tree} over actions (but not over transitions); that is, at every state we evaluate all possible actions, while state transitions are still obtained by sampling. This removes variance introduced by action-level sampling. Let $s'\qty(s, a)$ denote the successor state sampled when taking action $a$ in state $s$. Then \cref{eq:q_n_full} reduces to
\begin{equation} \label{eq:q_n}
    \hat{Q}^{(n)}\qty(s, a) = r\qty(s, a) + \gamma\max_{a' \in \mathcal{A}}\hat{Q}^{(n-1)}\qty(s'\qty(s, a), a'). \qquad \qty(n > 0)
\end{equation}
Also, we can further relax these targets into a forward-view Q($\lambda$) \citep{watkins1989qlearning,maei2010gqlambda,sutton2014qlambda} target, which is a continuous interpolation between the discrete $n$-step targets. 
Let $T$ the maximum horizon length and $t\qty(s)$ the iteration at which state $s$ occurred. Then the Q($\lambda$) target can be written as
\begin{equation} \label{eq:qlambda}
    \hat{Q}^{\lambda}\qty(s, a) = \qty(1 - \lambda)\sum_{n=1}^{T - t\qty(s) - 1}\lambda^{n-1}\hat{Q}^{(n)}\qty(s, a) + \lambda^{T - t\qty(s) - 1}\hat{Q}^{(T - t\qty(s))}\qty(s, a),
\end{equation}
where $\hat{Q}^{(0)}$ and $\hat{Q}^{(n)}$ $\qty(n > 0)$ follow the definitions of \cref{eq:q_0} and \cref{eq:q_n}. Now we can use $\hat{Q}^{\lambda}$ as an alternative to $\hat{Q}$ in \cref{eq:qlearning}. We call this resulting algorithm \textbf{full-width forward-view Q($\boldsymbol{\lambda}$)}.

\section{Stop-RAG}
\label{sec:methodology}
In this section, we describe Stop-RAG, our \emph{value-based} approach to decide whether further retrieval is necessary in an iterative RAG pipeline. We begin by formulating the iterative RAG process as an MDP, then apply full-width forward-view Q($\lambda$) described in \cref{sec:preliminaries} to derive an effective decision strategy. Our method naturally leverages offline data collected before training, without requiring additional online interaction.

\subsection{Iterative RAG as an MDP}
\label{subsec:mdp}

\begin{figure}[t!]
    \centering
    \subcaptionbox{Iterative RAG as an MDP (max iterations = 3)\label{fig:mdp}}{%
        \includegraphics[height=4.25cm]{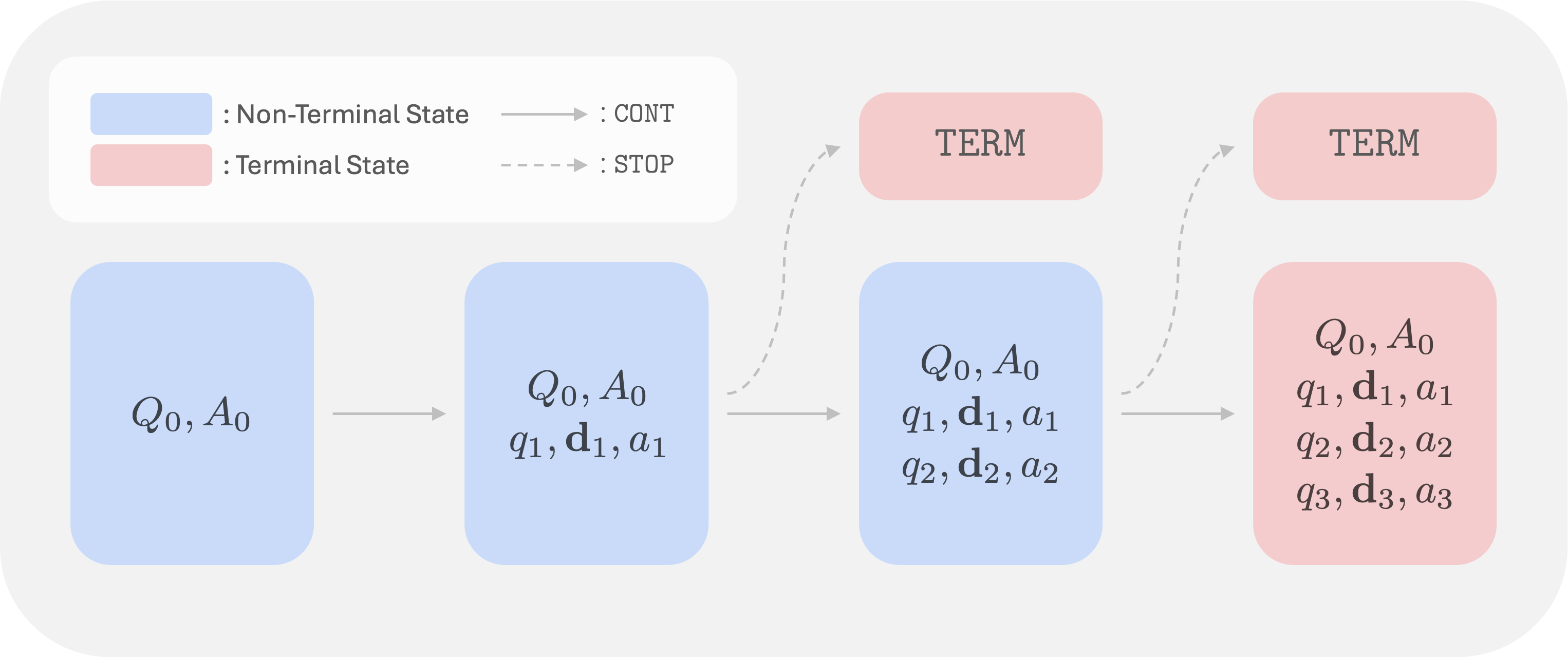}%
    }\hfill%
    \subcaptionbox{$\CONT$ in detail\label{fig:cont}}{%
        \includegraphics[height=4.25cm]{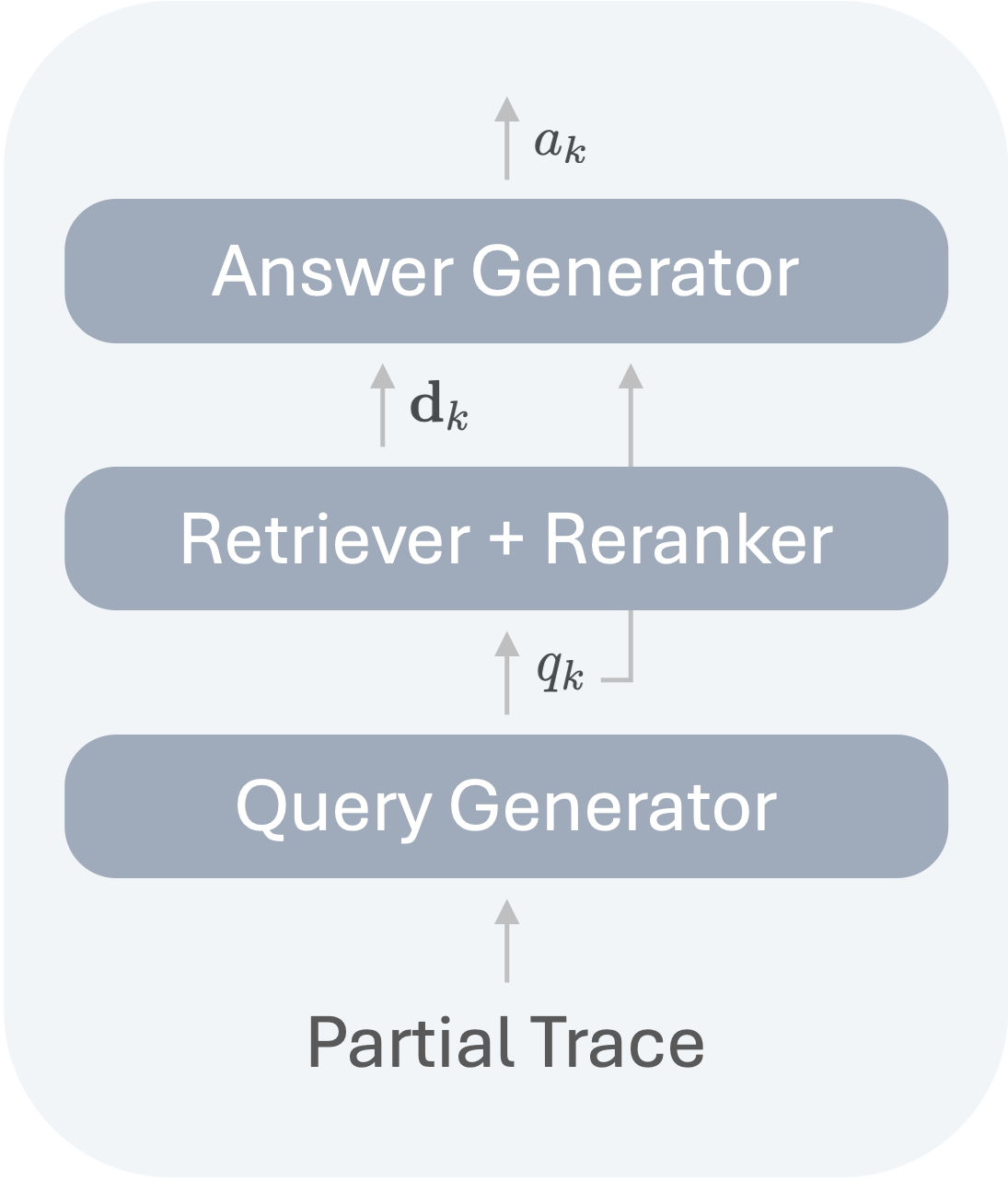}%
    }
    \caption{Illustration of the iterative RAG pipeline formulated as a finite-horizon MDP. Blue nodes denote non-terminal states while red nodes denote terminal states. At each step, the agent can either $\STOP$ (terminate to the absorbing state $\TERM$) or $\CONT$ (perform a RAG iteration). In addition, states that reach the maximum iteration limit are also treated as terminal. A RAG iteration consists of query generation ($q_k$), retrieval and reranking ($\vb{d}_k$), and intermediate answer generation ($a_k$). Rewards are given only at terminal states, reflecting the answer generation quality at that point.}
    \label{fig:iterative_rag}
\end{figure}

A typical iterative RAG pipeline can generally be illustrated as follows. Given a question $Q_0$ with ground-truth answer $A_0$, at each iteration of RAG we repeat a sequence of external interactions: (1) generate a query $q_k$ to perform retrieval; (2) retrieve, rerank, and select relevant documents $\vb{d}_k$; (3) generate an intermediate answer $a_k$ to the query using the retrieved documents. The process continues until the agent chooses to stop or reaches the maximum allowed number of iterations, at which point a final prediction $\hat{A}_0$ is produced based on all prior interaction records.

In many implementations, both query generation and intermediate answer generation are performed through LLM prompting \citep{corag, iterdrag}, often within a single inference call \citep{iterretgen}. If the stop decision is also prompt-based, it can likewise be integrated into the same call \citep{ircot,searcho1}.

This procedure can be modeled as a finite-horizon MDP, as illustrated in \Cref{fig:iterative_rag}. The set of prior interaction records at each iteration becomes a \emph{state}, i.e., the state at the $t$-th iteration is $\qty{Q_0, A_0} \cup \bigcup_{k=1}^{t}\qty{q_k, \vb{d}_k, a_k}$. We have two possible \emph{actions}: $\STOP$ and $\CONT$, which indicate whether to stop the pipeline or not, respectively. Any execution of $\STOP$ \emph{deterministically} transitions the system to an absorbing terminal state $\TERM$, while $\CONT$ invokes the actual RAG iteration, \emph{stochastically} transitioning to the next state. Note that other than $\TERM$, any state that has reached the maximum allowed number of iterations is also terminal. \emph{Rewards} are assigned only upon transitions to terminal states and reflect the quality of answer generation at that state.

\subsection{Stop-RAG training}
\label{subsec:training}

\paragraph{Specialization of Q($\boldsymbol{\lambda}$) to iterative RAG} Building on the MDP formulation of iterative RAG, we now return to the full-width forward-view Q($\lambda$) algorithm described in \cref{sec:preliminaries}. In particular, we specialize \cref{eq:qlambda} to our setting. Since we have a small action space ($\abs{\mathcal{A}} = 2$) and one action ($\STOP$) deterministically terminates the process, it is both tractable and natural to construct a full-width tree over actions. We assume no-discounting.

Let $T$ be the maximum number of iterations. For notational simplicity, denote the initial state $\qty{Q_0, A_0}$ as $s_0$, and define subsequent states recursively by
\begin{equation}
    s_{t+1} \coloneqq s'\qty(s_t, \CONT), \qquad t = 0, \ldots, T-1,
\end{equation}
under the assumption that the process does not stop prematurely. Note that $s'\qty(s_t, \STOP) = \TERM$ holds deterministically.

We now derive the Q($\lambda$) targets at an arbitrary state $s_t$. First, for $a = \STOP$, we have $\hat{Q}^{(n)}\qty(s_t, \STOP) = r\qty(s_t, \STOP)$ for $n = 1, \ldots, T-t$. Therefore \cref{eq:qlambda} simplifies to
\begin{equation}
    \hat{Q}^{\lambda}\qty(s_t, \STOP) = r\qty(s_t, \STOP).
\end{equation}

Next, for $a = \CONT$, unrolling \cref{eq:q_0,eq:q_n} yields
\begin{align}
    \hat{Q}^{(n)}\qty(s_t, \CONT) &= \max\biggl[\!\Bigl\{r\qty(s_{t+k}, \STOP)\Bigr\}_{k=1}^{n-1} \cup \Bigl\{\llb Q_\theta\qty(s_{t+n}, \STOP)\rrb, \llb Q_\theta\qty(s_{t+n}, \CONT)\rrb\Bigr\}\!\biggr], \label{eq:q_n_cont} \\
    \hat{Q}^{(T-t)}\qty(s_t, \CONT) &= \max\biggl[\!\Bigl\{r\qty(s_{t+k}, \STOP)\Bigr\}_{k=1}^{T-t-1} \cup \Bigl\{r\qty(s_{T-1}, \CONT)\Bigr\}\!\biggr],
\end{align}

where \cref{eq:q_n_cont} only applies for $n = 1, \ldots, T-t-1$. Then substituting into \cref{eq:qlambda} gives
\begin{equation} \label{eq:qlambda_cont}
    \hat{Q}^{\lambda}\qty(s_t, \CONT) = \qty(1 - \lambda)\sum_{n=1}^{T-t-1}\lambda^{n-1}\hat{Q}^{(n)}\qty(s_t, \CONT) + \lambda^{T-t-1}\hat{Q}^{(T-t)}\qty(s_t, \CONT).
\end{equation}
Especially, if $t = T-1$, this collapses to
\begin{equation}
    \hat{Q}^{\lambda}\qty(s_{T-1}, \CONT) = \hat{Q}^{(1)}\qty(s_{T-1}, \CONT) = r\qty(s_{T-1}, \CONT).
\end{equation}

\paragraph{Offline dataset construction} To run training on the Q($\lambda$) targets demonstrated above, we construct an offline dataset by generating full interaction trajectories and decomposing them into state-reward pairs. The procedure is as follows.
\begin{enumerate}[leftmargin=0.6cm]
    \item Begin with a corpus of questions $Q_0$ and corresponding ground-truth answers $A_0$.
    \item For each pair, we run the iterative RAG pipeline without intermediate stopping, always continuing up to the horizon limit $T$. This yields full-length trajectories containing all interaction records.
    \item Each prefix of a trajectory is treated as a partial trace, corresponding to a state in the MDP formulation. We include every such partial trace as a datapoint in the offline dataset. For each state, we estimate rewards by running multiple independent answer generation trials conditioned on that state, and scoring the outputs against $A_0$. Formally, for $t=1,\ldots,T-1$:
    \begin{align}
        r(s_t, \STOP) &= \frac{1}{N}\sum_{i=1}^{N}S\Bigl(\hat{A}_0^{(i)}\qty(s_t), A_0\Bigr), \\
        r(s_t, \CONT) &= \begin{cases}
            \frac{1}{N}\sum_{i=1}^{N}S\Bigl(\hat{A}_0^{(i)}\qty(s_T), A_0\Bigr), & \qty(t = T-1), \\
            0, & \qty(\text{otherwise}),
        \end{cases}
    \end{align}
    where $S$ is an arbitrary scoring function and $\hat{A}_0^{(i)}\qty(s)$ indicates the prediction generated from state $s$ at the $i$-th trial. In our experiments, we use F1 scores for $S$ and $N=8$.
    \item Filter out terminal states and states where $r\qty(s_t, \STOP) = 0$ and $\hat{Q}^{(T-t)}\qty(s_t, \CONT) = 0$. Discarding such states stabilizes training, since they provide no useful signal.
\end{enumerate}

This construction yields an offline dataset that spans all possible actions along each trajectory, while focusing training on informative signals. It enables consistent estimation of Q($\lambda$) targets without requiring online interaction.

\paragraph{Q($\boldsymbol{\lambda}$) training via function approximation} We train a function-approximated Q-network on the Q($\lambda$) targets described above. While the full-width formulation eliminates variance from action-level sampling, variance remains due to stochastic state transitions. To mitigate instability, we initialize training with $\lambda=1$, which corresponds to the longest-horizon targets and reduces bias at the cost of higher variance. As training progresses, we gradually decay $\lambda$ toward smaller values, thereby interpolating toward shorter-horizon backups that provide lower variance. This annealing schedule balances the bias-variance tradeoff inherent in Q($\lambda$), stabilizing learning while preserving the benefits of long-horizon targets. The complete training procedure is summarized in \Cref{alg:qlambda}.

\begin{algorithm}[t!]
\caption{Stop-RAG training with \emph{full-width forward view Q($\lambda$)}} \label{alg:qlambda}
\begingroup
\algrenewcommand{\algorithmicrequire}{\textbf{Initialize:}}%
\begin{algorithmic}[1]
\Require Q-network $Q_\theta$, $\lambda$, learning rate $\eta$
\State $\mathcal{D} \gets$ sample full trajectories w/o stopping, then split all into partial traces and calculate rewards
\While{not converged}
    \State sample batch $\mathcal{B} \subset \mathcal{D}$
    \State $\mathcal{L}_\theta \gets 0$
    \For{each state $s_t \in \mathcal{B}$} \Comment{run at batch-level}
        \State $\hat{Q}^\lambda\qty(s_t, \STOP) \gets r\qty(s_t, \STOP)$
        \If{$ t < T-1$}
            \State compute $Q_\theta\qty(s_{t+k}, \STOP)$ and $Q_\theta\qty(s_{t+k}, \CONT)$ for $k = 1, \ldots, T-t-1$ \Comment{no grad}
            \State $\hat{Q}^\lambda\qty(s_t, \CONT) \gets \qty(1 - \lambda)\sum_{n=1}^{T-t-1}\lambda^{n-1}\hat{Q}^{(n)}\qty(s_t, \CONT) + \lambda^{T-t-1}\hat{Q}^{(T-t)}\qty(s_t, \CONT)$
        \Else
            \State $\hat{Q}^\lambda\qty(s_t, \CONT) \gets r\qty(s_{T-1}, \CONT)$
        \EndIf
        \State $\mathcal{L}_\theta \gets \mathcal{L}_\theta + \qty[\qty(\hat{Q}^\lambda\qty(s_t, \STOP) - Q_\theta\qty(s_t, \STOP))^2 + \qty(\hat{Q}^\lambda\qty(s_t, \CONT) - Q_\theta\qty(s_t, \CONT))^2]$
    \EndFor
    \State take optimizer step on $\mathcal{L}_\theta$
    \State adjust $\lambda$ and $\eta$
\EndWhile
\end{algorithmic}
\endgroup
\end{algorithm}

\section{Experiments}
\label{sec:experiments}
\subsection{Evaluation setup}
\label{subsec:setup}

\paragraph{Datasets}
We evaluate Stop-RAG on three \emph{multi-hop QA} benchmarks: MuSiQue \citep{musique}, HotpotQA \citep{hotpotqa}, and 2WikiMultihopQA \citep{2wikimultihopqa}. These datasets require reasoning across multiple retrieval steps, with the number of steps varying by instance. This property makes them particularly suitable for evaluating our proposed framework. For training we use the official `train' partitions, and from the `dev' partitions we each randomly sample 1000 questions for validation and another 1000 for testing.

The retrieval corpus differs across datasets. For HotpotQA, we adopt the processed Wikipedia corpus officially released by the authors. For MuSiQue and 2WikiMultihopQA, which were originally formulated as reading comprehension tasks, we construct a retrieval corpus by aggregating all contexts provided in the `train', `dev', and `test' partitions, following \citet{ircot,adaptiverag}.

\paragraph{Evaluation metrics}
We report exact match (EM) and F1 scores between the generated prediction and gold answer, following \citet{squad}. We also provide the accuracy (Acc) score, which indicate whether the gold answer appears anywhere in the generated prediction, following \citet{popqa,toolformer}.

\subsection{Baselines}
\label{subsec:baselines}

Since Stop-RAG is designed as a plug-and-play stopping module, we evaluate it on top of two distinct iterative RAG pipelines. To cover both sides of the design space, we select one pipeline from prior work and construct another ourselves. CoRAG \citep{corag} serves as a representative method that involves generator fine-tuning and reports strong performance with a clear, reproducible setup, making it a suitable representative of this class. In contrast, pipelines without generator training vary substantially in prompts and retriever-generator configurations, and typically fall short in reported performance. To provide a fair and competitive counterpart in this category, we implement a modern, general pipeline that incorporates best practices in retrieval and prompting. For reference, we also include in \Cref{appendix:baselines} a table summarizing reported results of other existing pipelines.

For each pipeline, we compare three variants: the \textbf{raw} version, which uses a fixed number of iterations; the \textbf{LLM-Stop} version, where the model is prompted to decide when to stop; and the \textbf{Stop-RAG} version, which augments the pipeline with our proposed stopping mechanism. This setup allows us to evaluate not only the standalone effectiveness of Stop-RAG, but also its ability to improve performance over both fixed-iteration and naive prompting baselines.

Future work may explore applying Stop-RAG to other high-performing iterative RAG frameworks.

\subsection{Implementation details}
\label{subsec:implementation}

Our pipeline follows the general formulation of \cref{subsec:mdp}. Specifically, we use Llama-3.1-8B-Instruct \citep{llama3} as both the query and answer generator, Contriever-MSMARCO \citep{contriever} as the retriever, and bge-reranker-v2-m3 \citep{bge,bgem3} as the reranker. At each iteration 10 documents are retrieved and the top-ranked one is selected, with a maximum of 10 iterations allowed. For CoRAG, we reimplement the pipeline using the official checkpoint released by the authors and their original prompts, but replace the retriever and corpus with those used in our pipeline. The full set of prompts used in the experiments is provided in \Cref{appendix:prompts}.

For the Q-network in Stop-RAG, we adopt a DeBERTa-v3-large \citep{debertav3} backbone with two separate prediction heads---each a feed-forward network with one hidden layer---respectively corresponding to the $\STOP$ and $\CONT$ actions.
While Q-networks in general take both state and action as input, here the network conditions only on the state and predicts outputs for all actions, which is feasible given our small action space.
Each state $s_t$ is formatted as the main question $Q_0$ concatenated with all retrieved documents $\vb{d}_1, \ldots, \vb{d}_t$, separated by the separator token \texttt{[SEP]}. Note that intermediate queries and answers are not included in the state representation, as we found the retrieved evidence alone sufficient to learn effective stopping policies.
We provide the complete list of training hyperparameters in \Cref{appendix:hyperparameters}.

After training, we select the best checkpoint on the validation set. At inference time, rather than following the greedy policy induced by the trained Q-network, we compare the two head outputs and apply a margin-based decision rule: if $Q_\theta(s,\STOP) - Q_\theta(s,\CONT)$ exceeds a tuned threshold, the agent stops retrieval and produces a final answer; otherwise it continues retrieval. This threshold is chosen on the validation set.

\section{Results}
\label{sec:results}
\begin{table}[t!]
\centering
\caption{Performance comparison on MuSiQue, HotpotQA, and 2WikiMultihopQA. For each pipeline, we report results for the raw version which runs for a fixed number of iterations, a prompting-based stopping variant (LLM-Stop), and the same pipeline augmented with our method (Stop-RAG). Stop-RAG consistently improves performance over both fixed iteration and naive prompting, demonstrating its effectiveness as a plug-and-play stopping module.}
\label{tab:main_results}
\vskip 10pt
\begin{tabularx}{\textwidth}{l *{9}{>{\Centering\arraybackslash}X}}
\toprule
\multirow{2.5}{*}{Method} & \multicolumn{3}{c}{MuSiQue} & \multicolumn{3}{c}{HotpotQA} & \multicolumn{3}{c}{2WikiMultihopQA} \\
\cmidrule(lr){2-4} \cmidrule(lr){5-7} \cmidrule(lr){8-10}
& EM & F1 & Acc & EM & F1 & Acc & EM & F1 & Acc \\
\midrule
CoRAG ($L$ = 10, greedy) & 30.9 & 41.9 & 36.3 & \textbf{55.0} & \textbf{67.5} & \textbf{65.4} & 65.3 & 71.7 & 68.0 \\
\hspace{1mm} $\triangleright$ + LLM-Stop          & 31.1 & 42.0 & 36.3 & 54.9 & 67.6 & 64.8 & 65.4 & 72.2 & 67.8 \\
\hspace{1mm} $\triangleright$ + \textbf{Stop-RAG} & \textbf{31.5} & \textbf{43.0} & \textbf{36.9} & 54.7 & 67.4 & 64.4 & \textbf{65.7} & \textbf{72.5} & \textbf{68.2} \\
\midrule
Our Pipeline             & 34.5 & 44.8 & 41.1 & 51.0 & 65.0 & 60.8 & 64.9 & 73.1 & 65.9 \\
\hspace{1mm} $\triangleright$ + LLM-Stop          & 34.2 & 44.6 & 41.3 & 51.4 & 65.2 & 60.7 & 64.9 & 73.3 & 66.1 \\
\hspace{1mm} $\triangleright$ + \textbf{Stop-RAG} & \textbf{36.8} & \textbf{47.0} & \textbf{43.9} & \textbf{52.4} & \textbf{66.1} & \textbf{62.6} & \textbf{68.2} & \textbf{75.7} & \textbf{69.0} \\
\bottomrule
\end{tabularx}
\end{table}

\subsection{Main results}
\label{subsec:main_results}

\Cref{tab:main_results} presents the overall results of our experiments.
Within our pipeline, Stop-RAG consistently outperforms both baselines across all benchmarks. This indicates that naive prompting is insufficient for effective stopping and highlights the importance of a principled stopping mechanism. When applied to CoRAG, Stop-RAG demonstrates its effectiveness as a plug-and-play module. On MuSiQue and 2WikiMultihopQA, applying Stop-RAG achieves higher performance than both baselines, showing that the method generalizes across different iterative RAG architectures. These results emphasize that Stop-RAG is not tied to a specific retriever-generator design, but can integrate seamlessly into varied systems to yield consistent gains.

However, on HotpotQA, Stop-RAG slightly underperforms both baselines. This result likely stems from the fact that CoRAG employs a fine-tuned generator trained under the assumption of a fixed number of iterations. This training setup appears to make the generator more robust to distractors and less sensitive to retrieval inefficiency, which may reduce the relative benefit of adaptive stopping. Similar behavior can be observed with CoRAG on other datasets, where Stop-RAG yields only slight improvements compared to its effect on our pipeline. In HotpotQA specifically, where the retrieval corpus is broader and more challenging than the other benchmarks, retrieval recall plays a critical role. Thus, continuing retrieval for the full iteration count appears to outweigh the cost of additional distractors, which may explain the stronger performance of the fixed-iteration baseline.

Overall, these results highlight two key insights: (1) Stop-RAG substantially improves pipelines by identifying more optimal stopping points, and (2) the method demonstrates broad applicability as a plug-and-play component across different architectures.

\subsection{Analysis}
\label{subsec:analysis}

\begin{table}[t!]
\parbox{.63\linewidth}{
\centering
\caption{Retrieval performance on MuSiQue and 2WikiMultihopQA, tested on our pipeline variants. Precision and recall are measured for document retrieval, while `Steps' denotes the average number of iterations executed. Compared to LLM-Stop, Stop-RAG runs slightly longer on average, which increases recall and leads to better performance.}
\label{tab:retrieval_results}
\vskip 10pt
\setlength{\tabcolsep}{2pt}
\begin{tabularx}{\linewidth}{l *{6}{>{\Centering\arraybackslash}X}}
\toprule
\multirow{2.5}{*}{Method} & \multicolumn{3}{c}{MuSiQue} & \multicolumn{3}{c}{2WikiMHQA} \\
\cmidrule(lr){2-4} \cmidrule(lr){5-7}
& Prec & Recall & Steps & Prec & Recall & Steps \\
\midrule
Our Pipeline                                         & 17.4 & 68.2 & 10.0 & 21.4 & 88.3 & 10.0 \\
\hspace{0.05mm} $\triangleright$ + LLM-Stop          & 31.6 & 64.5 & 7.6  & 55.0 & 79.1 & 5.0 \\
\hspace{0.05mm} $\triangleright$ + \textbf{Stop-RAG} & 22.2 & 66.5 & 8.6  & 49.5 & 82.6 & 5.1 \\
\bottomrule
\end{tabularx}
}
\hfill
\parbox{.35\linewidth}{
\centering
\caption{Performance of Stop-RAG variants on MuSiQue when using different learning targets. Detailed descriptions of each method are provided in \Cref{appendix:ablation}.}
\label{tab:ablation}
\vskip 10pt
\setlength{\tabcolsep}{3pt}
\begin{tabularx}{\linewidth}{l *{3}{>{\Centering\arraybackslash}X}}
\toprule
\multirow{2.5}{*}{Method} & \multicolumn{3}{c}{MuSiQue} \\
\cmidrule(lr){2-4}
& EM & F1 & Acc \\
\midrule
Binary                             & 33.3 & 43.7 & 40.3 \\
MC                                 & 36.0 & 46.2 & 43.0 \\
\textbf{Q($\boldsymbol{\lambda}$)} & \textbf{36.8} & \textbf{47.0} & \textbf{43.9} \\
Q(0)                               & 36.0 & 46.3 & 42.8 \\
\bottomrule
\end{tabularx}
}
\end{table}

\paragraph{Retrieval results}
To further assess the effect Stop-RAG, we analyze retrieval performance on MuSiQue and 2WikiMultihopQA, where the retrieval corpus is manually constructed and ground-truth supporting documents are given. As shown in \Cref{tab:retrieval_results}, Stop-RAG executes slightly more steps on average than LLM-Stop, resulting in higher recall. This suggests that Stop-RAG improves end-task performance by adaptively continuing retrieval when additional evidence is beneficial, rather than terminating early like LLM-Stop does. In doing so, it achieves a better balance between retrieval depth and precision.

\paragraph{Ablation study}
To better understand the impact of our learning objective, we performed an ablation study comparing several alternatives to our Q($\lambda$)-based objective. In addition to our main approach with annealed Q($\lambda$) targets, which balance variance reduction with forward-looking credit assignment, we evaluate Monte Carlo (MC) returns (no bootstrapping) and one-step temporal-difference (TD) targets (fully bootstrapped). For clarity, we refer to the one-step TD variant as Q(0), by analogy to TD(0). We also test a binary classification variant, where each state is labeled positive or negative depending on whether the MC return for $\STOP$ exceeds that for $\CONT$. Full details of the training objectives are provided in \Cref{appendix:ablation}.

As shown in \Cref{tab:ablation}, Q($\lambda$) achieves the best overall balance, yielding the highest F1 and accuracy. MC performs reasonably well but suffers from high variance, while Q(0) also performs well but is limited by its myopic nature. The binary variant lags behind, suggesting that richer value-based targets are crucial for learning effective stopping policies.

\section{Conclusion}
\label{sec:conclusion}
We introduced Stop-RAG, a value-based controller for adaptive stopping in iterative retrieval-augmented generation. By framing the problem as a finite-horizon MDP and training with Q($\lambda$) targets, Stop-RAG provides forward-looking estimates of the benefits of continued retrieval, enabling more optimal stopping decisions than fixed-iteration or prompting-based baselines. 

Experiments across multiple multi-hop QA benchmarks show that Stop-RAG consistently improves performance, demonstrating its ability to balance retrieval recall and precision while remaining modular and compatible with black-box LLMs. These results demonstrate the effectiveness of value-based stopping control for iterative RAG and highlight the potential for adaptive decision making in broader LLM-driven systems.

{\footnotesize
\bibliographystyle{plainnat_with_preprint}
\bibliography{neurips_2025}}

\newpage

\appendix
\section{Performance summary of existing iterative RAG pipelines}
\label{appendix:baselines}

\Cref{tab:baseline_results} summarizes reported results of prior iterative RAG methods for reference. We include this table primarily to motivate the design of our own pipeline: while several iterative RAG systems exist, their implementations vary substantially in prompts, retrievers, and generators, and reported results are inconsistent across datasets. Scores for Iter-RetGen, Adaptive-RAG, Probing-RAG, IterDRAG, and Search-o1 are taken from their original papers. Results for IRCoT come from \citet{adaptiverag} and results for Self-RAG from \citet{corag}, since the original works do not report full results on these datasets. When multiple configurations are reported, we select the setup most comparable to ours, while reporting lower-spec results if they outperform higher-spec variants. The implementation details of each pipeline are as follows.
\begin{itemize}[leftmargin=0.6cm]
    \item \textbf{IRCoT} and \textbf{Adaptive-RAG}: Uses FLAN-T5-XL (3B) and FLAN-T5-XXL (11B) \citep{flant5} as the generator and BM25 \citep{bm25} as the retriever.
    \item \textbf{Iter-RetGen}: Uses text-davinci-003 \citep{instructgpt} as the generator and Contriever-MSMARCO as the retriever.
    \item \textbf{Self-RAG}: Uses the same fine-tuned Llama-2-7B \citep{llama2} checkpoint provided by the original authors but replaces the retrieval system with E5-large \citep{e5}.
    \item \textbf{IterDRAG}: Uses Gemini 1.5 Flash \citep{gemini1.5} as the generator and Gecko 1B \citep{gecko} as the retriever.
    \item \textbf{Search-o1}: Uses QwQ-32B \citep{qwen2.5,qwq32b} as the generator and the Bing Web Search API as the retriever.
\end{itemize}

\begin{table}[b!]
\centering
\caption{Reported performance of existing iterative RAG methods on MuSiQue, HotpotQA, and 2WikiMultihopQA. Gray-highlighted text denotes results from lower-spec models that outperform larger variants. Missing results (`--') indicate those not reported in the corresponding sources.}
\label{tab:baseline_results}
\vskip 10pt
\begin{tabularx}{\textwidth}{l *{9}{>{\Centering\arraybackslash}X}}
\toprule
\multirow{2.5}{*}{Method} & \multicolumn{3}{c}{MuSiQue} & \multicolumn{3}{c}{HotpotQA} & \multicolumn{3}{c}{2WikiMultihopQA} \\
\cmidrule(lr){2-4} \cmidrule(lr){5-7} \cmidrule(lr){8-10}
& EM & F1 & Acc & EM & F1 & Acc & EM & F1 & Acc \\
\midrule
IRCoT (\textcolor{gray}{3B} \& 11B) & \textcolor{gray}{23.0} & \textcolor{gray}{31.9} & \textcolor{gray}{25.8} & 47.0 & 57.8 & 49.4 & 57.6 & 67.7 & 64.0 \\
Iter-RetGen ($T$ = 7)       & 26.1 & 42.0 & -- & 45.1 & 60.4 & - & 35.4 & 47.4 & -- \\
Self-RAG (7B)               & 4.6 & 13.2 & -- & 16.6 & 29.4 & -- & 12.2 & 24.1 & -- \\
Adaptive-RAG (\textcolor{gray}{3B} \& 11B) & \textcolor{gray}{23.6} & \textcolor{gray}{31.8} & \textcolor{gray}{26.0} & 44.2 & 54.8 & 46.8 & 47.6 & 57.4 & 54.0 \\
IterDRAG (32k)              & 12.5 & 23.1 & 19.7 & 38.3 & 49.8 & 44.4 & 44.3 & 54.6 & 56.8 \\
Search-o1                   & 16.6 & 28.2 & -- & 45.2 & 57.3 & -- & 58.0 & 71.4 & -- \\
\bottomrule
\end{tabularx}
\end{table}

\section{Hyperparameters}
\label{appendix:hyperparameters}

We train the Stop-RAG Q-network with standard settings. The full list is provided in \Cref{tab:hyperparameters}.

\begin{table}[b!]
\centering
\caption{Hyperparameters used for training the Stop-RAG Q-network.}
\label{tab:hyperparameters}
\vskip 10pt
\begin{tabular}{ll}
\toprule
Hyperparamter & Value \\
\midrule
Hidden size of prediction heads & 4096 \\
Optimizer & AdamW \citep{adamw} \\
Learning rate & Cosine decay from 5e-5 to 0.0 \\
Warmup ratio & 0.1 \\
Weight decay & 0.01 \\
Batch size & 128 \\
Epochs & 3 for MuSiQue, 1 for HotpotQA and 2WikiMultihopQA \\
Precision & Mixed precision (bfloat16) \\
$\lambda$ & Cosine decay from 1.0 to 0.1 \\
\bottomrule
\end{tabular}
\end{table}

\section{Training objectives of ablation variants}
\label{appendix:ablation}
Below we detail the training objectives considered in our ablation study.

\paragraph{Monte Carlo (MC) target}
A straightforward option is to use Monte Carlo (MC) rollouts of the stopping process. The Q-value is estimated as the immediate reward plus the maximum return achievable from subsequent states:
\begin{equation}
    \hat{Q}^{\text{MC}}\qty(s, a) = r\qty(s, a) + \gamma\max_{a' \in \mathcal{A}}\hat{Q}^{\text{MC}}\qty(s'\qty(s, a), a').
\end{equation}
For our MDP formulation, we have the following.
\begin{align}
    \hat{Q}^{\text{MC}}\qty(s_t, \STOP) &= r\qty(s_t, \STOP), \\
    \hat{Q}^{\text{MC}}\qty(s_t, \CONT) &= \max\biggl[\Bigl\{r\qty(s_{t+k}, \STOP)\Bigr\}_{k=1}^{T-t-1} \cup \Bigl\{r\qty(s_{T-1}, \CONT)\Bigr\}\biggr].
\end{align}
The training objective here is the same as in our original setup; only the target differs.

\paragraph{Q(0) target}
We also consider the standard one-step temporal difference target, which is equivalent to the $\lambda=0$ fixed case of Q($\lambda$):
\begin{equation}
    \hat{Q}^{\text{Q(0)}}\qty(s, a) = \hat{Q}^{\lambda}\qty(s, a) \Big\vert_{\lambda = 0} = \hat{Q}^{(1)}\qty(s, a) = r\qty(s, a) + \gamma\max_{a' \in \mathcal{A}} \llb Q_\theta\qty(s'\qty(s, a), a') \rrb.
\end{equation}
For our MDP formulation, we have the following.
\begin{align}
    \hat{Q}^{\text{Q(0)}}\qty(s_t, \STOP) &= r\qty(s_t, \STOP), \\
    \hat{Q}^{\text{Q(0)}}\qty(s_t, \CONT) &= \max\Bigl\{Q_\theta\qty(s_{t+1}, \STOP), Q_\theta\qty(s_{t+1}, \CONT)\Bigr\}.
\end{align}
As with the MC target, the training objective is unchanged; only the target values are replaced.

\paragraph{Binary classification}
Finally, we test a binary formulation where the model directly predicts whether to stop or continue at each state. The label is defined by comparing the Monte Carlo estimates of stop and continue:
\begin{equation}
    y\qty(s_t) = \mathds{1}_{\hat{Q}^{\text{MC}}\qty(s_t, \STOP) \geq \hat{Q}^{\text{MC}}\qty(s_t, \CONT)}.
\end{equation}
The training objective uses the standard binary cross entropy loss:
\begin{equation}
    \mathcal{L}_\theta\qty(s_t) = -y\qty(s_t)\log\sigma\qty(\mathcal{M}_\theta\qty(s_t)) - \bigl(1 - y\qty(s_t)\bigr)\log\bigl(1 - \sigma\qty(\mathcal{M}_\theta\qty(s_t))\bigr),
\end{equation}
where $\mathcal{M}_\theta$ is the stopping model parameterized by $\theta$, which we optimize.

\section{Prompts}
\label{appendix:prompts}
This section presents the key prompts used in our experimental setup.
\Cref{prompt:qg,prompt:iag,prompt:sd} correspond to those used in the \textbf{query generation}, \textbf{intermediate answer generation}, and \textbf{LLM-Stop} modules, respectively. 
Blue-highlighted text in brackets indicates placeholders to be filled in. 

In particular, \Cref{prompt:sd} implements the stopping mechanism for the \textbf{LLM-Stop} baseline described in \cref{subsec:baselines}, enabling the generator itself to decide whether to continue retrieval or stop.

For final answer generation, we employ IRCoT \citep{ircot}-style prompting, following the prompt template and in context examples of the original work.

\begin{prompt}[label=prompt:qg]{Query Generation}
\setlength{\parskip}{8pt}
\textbf{System Prompt:}

Given an original question and a series of follow-up questions and answers, generate the next logical follow-up question that targets a specific missing piece of information needed to answer the original question.

The question will be used to retrieve additional information from a knowledge base (e.g., Wikipedia).

Process:\\
- You receive:\\
\hspace*{1em}- Main question: <original question>  \\
\hspace*{1em}- Zero or more rounds of:\\
\hspace*{2em}- Follow up: <previous follow up question>  \\
\hspace*{2em}- Document: <top Wikipedia snippet>  \\
\hspace*{2em}- Intermediate answer: <answer to the Follow-up question based on the Document>

- Your task:\\
\hspace*{1em}1. Ask the next logical follow-up question.\\
\hspace*{1em}2. Base it solely on gaps in the existing trace.\\
\hspace*{1em}3. Do not repeat information already covered.\\
\hspace*{1em}4. Make sure it targets exactly the missing fact you need to answer the original question.\\
\hspace*{1em}5. Output only the new question itself; no explanations or extra text.

Here are some examples:\\
\textcolor{blue}{\texttt{[Few-shot examples of the query generation process]}}

\vskip 10pt

\textbf{User Prompt:}

Main question: \textcolor{blue}{\texttt{[Main question]}} \\
\textcolor{blue}{\texttt{[Current trace]}}

Respond with a simple follow-up question that will help answer the main question, do not explain yourself or output anything else.
\end{prompt}

\begin{prompt}[label=prompt:iag]{Intermediate Answer Generation}
\setlength{\parskip}{8pt}
\textbf{System Prompt:}

Given a question with its corresponding Wikipedia snippet, generate an answer using only the provided snippet.

Process:\\
- You receive:\\
\hspace*{1em}- Question: <a question>  \\
\hspace*{1em}- Document: <Wikipedia snippet for the question>

- Your task:\\
\hspace*{1em}1. Focus on answering the question.\\
\hspace*{1em}2. Use only the provided Document as your evidence.\\
\hspace*{1em}3. Output only the answer, with no additional text.\\
\hspace*{1em}4. Keep the answer concise and directly relevant to the question.

Here are some examples:\\
\textcolor{blue}{\texttt{[Few-shot examples of the intermediate answer generation process]}}

\vskip 10pt

\textbf{User Prompt:}

\textcolor{blue}{\texttt{[Current query and retrieved documents]}}
\end{prompt}

\begin{prompt}[label=prompt:sd]{LLM-Stop}
\setlength{\parskip}{8pt}
\textbf{System Prompt:}

Given an original question and a series of follow-up questions each paired with its top Wikipedia snippet plus intermediate answers, analyze step by step whether there is sufficient information to provide a complete and accurate final answer to the original question.

Process:\\
- You receive:\\
\hspace*{1em}- Main question: <original question>  \\
\hspace*{1em}- One or more rounds of:\\
\hspace*{2em}- Follow up: <follow-up question>  \\
\hspace*{2em}- Document: <top Wikipedia snippet>  \\
\hspace*{2em}- Intermediate answer: <answer to the Follow-up question based on the Document>

- Your task:\\
\hspace*{1em}1. Think step by step about what information is needed to answer the original question.\\
\hspace*{1em}2. Analyze what information has been gathered so far.\\
\hspace*{1em}3. Determine if any critical information is still missing.\\
\hspace*{1em}4. Make a reasoned decision about whether to stop or continue.

Format your response as:\\
Analysis: Step-by-step reasoning about information completeness\\
Decision: <STOP> or <CONTINUE>

Decision criteria:\\
- <STOP>: All components of the original question can be answered with current information such that a complete and final answer can be derived from it.\\
- <CONTINUE>: Missing any required information needed to answer the original question.

Here are some examples:\\
\textcolor{blue}{\texttt{[Few-shot examples of the LLM-Stop process]}}

\vskip 10pt

\textbf{User Prompt:}

Main question: \textcolor{blue}{\texttt{[Main question]}}\\
\textcolor{blue}{\texttt{[Current trace]}}

Based on the information provided, give a short analysis of whether the original question can be answered. Then, decide whether to stop or continue gathering information.
\end{prompt}

\end{document}